\documentclass[sn-mathphys-num]{sn-jnl}

\usepackage{graphicx}%
\usepackage{multirow}%
\usepackage{amsmath,amssymb,amsfonts}%
\usepackage{amsthm}%
\usepackage{mathrsfs}%
\usepackage[title]{appendix}%
\usepackage{xcolor}%
\usepackage{textcomp}%
\usepackage{manyfoot}%
\usepackage{booktabs}%
\usepackage[ruled, vlined]{algorithm2e}
\usepackage{algorithmic}
\usepackage{listings}%
\usepackage{rotating}    
\usepackage{caption}     
\usepackage{float}       
\usepackage{lineno}

\definecolor{red}{rgb}{1.0, 0.0, 0.0}

\RequirePackage{amsthm}

\theoremstyle{thmstyleone}%
\theoremstyle{thmstyletwo}%

\theoremstyle{thmstylethree}%

\begin{document}

\title[Article Title]{GMD: Gaussian Mixture Descriptor for Pair Matching of 3D Fragments}


\author[1]{\fnm{Meijun} \sur{Xiong}}\email{1875674254@qq.com}
\equalcont{These authors contributed equally to this work.}
\author[1]{\fnm{Zhenguo} \sur{Shi}}\email{2673720479@qq.com}
\equalcont{These authors contributed equally to this work.}

\author[1]{\fnm{Xinyu} \sur{Zhou}}\email{xinyuzhou77@163.com}

\author*[1]{\fnm{Yuhe} \sur{Zhang}}\email{zhangyuhe0601@nwu.edu.cn}

\author*[1]{\fnm{Shunli} \sur{Zhang}}\email{slzhang@nwu.edu.cn}

\affil[1]{\orgdiv{School of Information Science and Technology}, \orgname{Northwest University}, \orgaddress{\street{Xuefu Road, No. 1, Changan District}, \city{Xi’an City}, \postcode{710127}, \state{State}, \country{China}}}


\abstract{In the automatic reassembly of fragments acquired using laser scanners to reconstruct objects, a crucial step is the matching of fractured surfaces. In this paper, we propose a novel local descriptor that uses the Gaussian Mixture Model (GMM) to fit the distribution of points, allowing for the description and matching of fractured surfaces of fragments. Our method involves dividing a local surface patch into concave and convex regions for estimating the $k$ value of GMM. Then the final Gaussian Mixture Descriptor (GMD) of the fractured surface is formed by merging the regional GMDs. To measure the similarities between GMDs for determining adjacent fragments, we employ the $L_2$ distance and align the fragments using Random Sample Consensus (RANSAC) and Iterative Closest Point (ICP). The extensive experiments on real-scanned public datasets and Terracotta datasets demonstrate the effectiveness of our approach; furthermore, the comparisons with several existing methods also validate the advantage of the proposed method.}




\maketitle

\section{Introduction}\label{sec1}

Computer-aided techniques have become increasingly important in various fields, due to their ability to provide virtual representation and automate the reconstruction of 3D objects from laser-scanned fragments. The process of reconstructing broken objects often requires matching the shapes of the fractured surfaces. However, the exposed break-surfaces of cultural relics like ceramic artifacts are always defectively geometric models \cite{DIANGELO202210} with disorder, irregular and incomplete shapes, rendering it more challenging to reassemble the fragments. Furthermore, the lack of color or geometric texture associated with the original surfaces (the surfaces of the cultural relics), for example, the fragments of Terracotta Warriors and Horses, extremely increases the difficulty of matching the fractured surfaces. Additionally, it is often unclear whether the fractured surfaces being matched are perfectly or partially matched. Partial matching (which frequently occurs on fragments of cultural relics) means a notably lower overlapping ratio between two fractured surfaces to be matched, directly adding to the difficulty of finding real correspondences on a large surface. 

Dealing with the above challenges, it will be more accurate and efficient if we first sample surface patches on the fractured surface and then describe and match the sampled surface patches, as identified in most previous methods \cite{8938803,Son2017ReassemblyOF,10.1145/3417711}. Therefore, describing and matching the surface patches are crucial steps for this task. However, the existing numerous surface descriptors, such as Fast Point Feature Histograms (FPFH) \cite{rusu2009fast}, 
Signature of Histograms of Orientations (SHOT) \cite{2014SHOT}, and Spin Image \cite{assfalg2007content}, consider the surface is complete, focusing solely on extracting more information to guarantee an accurate similarity measurement. However, this assumption overlooks the challenges posed by fractured surfaces with defective parts, which highly decreases the shape similarities of the local surface patches around the real corresponding point pairs. 

In this work, we design a surface descriptor based on the Gaussian Mixture Model (GMM), namely the Gaussian Mixture Descriptor (GMD) for describing surface patches. GMM has been widely used in various studies due to its ability to approximate any point distribution \cite{dempster1977maximum}. GMM is a statistical model that can extract a high-level abstract of the surface patch, instead of focusing on the detailed information of the surface patch, making it more suitable for handling surface patches with defective parts. 

GMM has proven to be effective in fitting data distribution, geometric information, or convex-hull of point clouds \cite{FAN2016126,5674050,liu2021lsg,qu2021point,yang2022robust}. However, these methods do not estimate the parameters of GMM over a triangular mesh, but rather directly use some values of a reference triangular facet to form a GMM \cite{FAN2016126}. The generalization of these methods is relatively limited as they highly depend on the quality of the reconstructed triangular meshes. Furthermore, these approaches mostly focus on global transformations of the entire point clouds, 
making them more suitable for objects with obvious sharp and smooth geometric structures but less effective in handling fractured surfaces, which are likely to be a slice.

Unlike previous GMM-based methods like \cite{FAN2016126} and \cite{yang2022robust}, we propose a new method to build a local descriptor based on GMM, forming a descriptor GMD for local surface patches on fractured surfaces to 
preserve the local neighborhood information.
The initial step in our approach involves utilizing the Scale Invariant Feature Transform (SIFT) method implemented in the Point Cloud Library (PCL) \cite{sift}
to sample feature points from the fractured surface, and then compute the Local Reference Frame (LRF) at each feature point for building the descriptors. 
Next, we calculate the GMDs for both concave and convex regions on the fractured surfaces respectively, 
using the angle between the normal vector of a point and the fitted plane of neighboring points within a certain radius to distinguish concave and convex regions. 
Then, we estimate the parameters of GMD using the Expectation-Maximization (EM) algorithm \cite{Embretson2000ItemRT} to fit the point distribution in concave and convex regions, forming the regional GMDs for concave and convex regions.
Finally, we merge the regional GMDs into one GMD to describe the shape of the local surface patches. 
The similarities between GMDs are utilized to determine the matching of fragments, 
and RANSAC and ICP are employed to align the matching fragments. 

To evaluate the proposed method, we conducted experiments on various real-world datasets and compared our method with several representative descriptors such as FPFH \cite{rusu2009fast}, 
SHOT \cite{2014SHOT}, Spin Image \cite{assfalg2007content}, 
and some recently proposed methods like TEASER \cite{Yang20tro-teaser} and GROR \cite{GROR}. We also used six different evaluation metrics for quantitative analysis and presented several qualitative results. The results indicate that the proposed GMD outperforms the other descriptors and demonstrates its effectiveness and accuracy, 
and the proposed method has been experimentally shown to be effective in reassembling real-world objects.

The main contributions of this work are as follows:

(1) A novel surface patch descriptor namely GMD is proposed.

(2) Six evaluation metrics including the Percentage of Coverage (PoC), Angle of the normal vector of the planes(AoNV), Angle of the normal vector of the local patch(localAoNV), Max Angle (MaA), Min Angle (MiA), and Mean Angle (MeA) between the normal vectors of the point pairs, are proposed to evaluate the performance of reassembly algorithms.

\section{Related work}\label{sec2}
\subsection{GMM on point clouds}\label{subsec2}

GMM has been successfully applied in various fields, such as image segmentation \cite{info13020098}, background subtraction \cite{mahajan2022detection}, 
and speaker verification \cite{AVILA202121}. In point cloud registration, GMM has also been widely used to fit a point set. 
For instance, inspired by the idea of minimizing the $L_{2}$ distance between Gaussian mixtures, 
Jian \textit{et al.} \cite{5674050} presented a registration framework featuring a closed-form expression that enabled an efficient registration algorithm. 
Liu \textit{et al.} \cite{liu2021lsg} presented a new approach termed CPD with Local Surface Geometry (LSG-CPD) for performing rigid registration of point clouds. 
Qu and Lee \cite{qu2021point} modeled the entire point set as a Gaussian mixture model and converted the registration problem of two point sets into a minimization problem of the improved KL divergence between two GMMs. 
Yang \textit{et al.} \cite{yang2022robust} presented a non-rigid point set registration method that deals with data containing various types of degradation. 
The authors formulated the registration problem as a density estimation problem based on a GMM. 
Fan\textit{ et al}. \cite{FAN2016126} presented a technique for calculating a weighted GMM over each point set's convex hull. 
To improve the method's performance, three conditions, proximity, area conservation, and projection consistency are considered in the model.

\subsection{Fragment reassembly}\label{subsec3}
Over recent years, several work have discussed the advances in the technologies for reconstructing fragments.\par
The template guidance-based methods use distribution information, axisymmetric information, 
and rotationally symmetric information on surfaces as a priori to guide the reassembly of fragments \cite{Zhang20153DFR}. 
Statistical distribution and symmetric information-based methods rely on a complete model as a template or a pre-built 
template library to determine the position of a fragment in the global context. 
The fragment is matched against the template, and upon the successful match, the fragments are aligned together to achieve the goal of reassembling. 
These approaches have been studied in detail in several previous works \cite{9022556, 10.1109/CVPR.2013.40}.\par
Various methods based on curves matching \cite{wang2022reconstructing} have been proposed for fragment reassembly.
Edge extraction techniques are always utilized to obtain the break curves of broken objects, 
and the similarity of these curves is then measured using certain rules \cite{10.1145/3460393, ZHANG2018191}. 
These methods are primarily applied in the matching two-dimensional images or the reassembly of fragments containing thin fractured surfaces, 
such as frescoes and ceramics \cite{Tsiafaki2016VirtualRA, article1}. However, these approaches are limited in their ability to handle more complex and irregularly shaped 
fractured surfaces, which are commonly encountered in 3D point clouds.\par
The fractured surface-based approach is accomplished by measuring the similarity of fractured surfaces. 
Savelonas \textit{et al.} \cite{Savelonas2017ExploitingUS} used the number of feature curves and the geometric texture of surfaces, 
along with a heuristic test to evaluate the rationality of surface pairs matching. 
Wu and Wang \cite{WU201855} utilized the Hausdorff distance and the modified 4-point congruent set algorithms to identify potential simple chordless cycles 
for matching fractured sand particles. Li and Geng \cite{8938803} presented a simple pairwise matching method for 3D fragment reassembly 
that uses boundary curves and concave-convex patches to accelerate and optimize the matching. 
Cakir \textit{et al.} \cite{9559005} proposed a Minimum Bounding Box (MBB) based alignment and key-point-based matching method. 
Paulano \textit{et al.} \cite{PAULANOGODINO201793} developed an automatic method that calculates the contact region between two bone fragments point clouds generated from CT images. 
Son T \textit{et al.} \cite{Son2017ReassemblyOF} proposed a descriptor based on the convex/concave information of a point on the fractured surfaces to describe the fractured surface 
and to find the counterpart fractured surface effectively. 
Wang\textit{ et al.} \cite{10.1145/3417711} defined a probability-based method to automatically reassemble a large collection of fragments of unknown geometric shapes, 
using Link-Chain Descriptors (LCD) and Spatial Distribution Descriptors (SDD) to detect and describe the matching objects. 
These methods have been applied to reassemble two-dimensional images or three-dimensional fractured surfaces.

Many deep learning-based fragment reassembly methods have also emerged. MatchMakerNet \cite{10350868} utilized graph convolution alongside a simplified version of DGCNN to automate the pairing of object fragments for reassembly. Chen\textit{et al.} \cite{chen2022neural} proposed Neural Shape Mating (NSM) to tackle pairwise 3D geometric shape alignment. Sellán \textit{ et al.} \cite{2022breaking} introduced Breaking Bad, a large-scale dataset of fractured objects that consists of over one million fractured objects simulated from ten thousand base models. The reported results highlight the high complexity of this task, given that synthetically generated fragments devoid of physical deterioration were only roughly aligned \cite{2022breaking}.

\section{Preliminaries}\label{sec3}
GMM is a parametric probability density function that can be represented as a weighted sum of Gaussian component densities \cite{NIPS1999_97d98119}. EM algorithm \cite{Embretson2000ItemRT} is generally used for calculating the parameters in GMM, thereby forming the model.

In particular, given random variables \textit{X}(\textit{D} dimensions), the GMM can be expressed using Equ.(\ref{equ1}):
\begin{equation}
p(x)=\sum_{i=1}^{k} \omega_{i} \phi\left(x \mid \mu_{i}, \Sigma_{\mathrm{i}}\right) \label{equ1}
\end{equation}
where
\begin{equation}
\phi\left(x \mid \mu_{i}, \Sigma_{i}\right)=\frac{\exp \left[-\frac{1}{2}\left(x-\mu_{i}\right)^{T} \sum_{i}^{-1}\left(x-\mu_{i}\right)\right]}{\sqrt{(2 \pi)^{D}\left|\operatorname{det}\left(\Sigma_{i}\right)\right|}} \label{equ2}
\end{equation}
where $x$ represents the samples, $k$ is the number of Gaussian components,   
$\omega_{i}$ is the weight of $i$th Gaussian components, $\mu_{i}$ is the mean vector of $i$th (0 < $i$ < $k$) Gaussian components, 
$\Sigma_{i}$ is the covariance matrix.

The EM algorithm \cite{Embretson2000ItemRT} usually has two steps: 

1) E step: calculate the posterior probability $\gamma(z_nk)$ based on the current $\omega_{k}$, $\mu_{k}$, $\sigma_{k}$ :
\begin{equation}
      \gamma(z_nk)=\frac{\omega_k\mathcal{N}(\boldsymbol{x}_{\mathrm{n}}|\boldsymbol{\mu}_{\mathrm{n}},\sum_n)}{\sum_{j=1}^K\omega_k\mathcal{N}(\boldsymbol{x}_{\mathrm{n}}|\boldsymbol{\mu}_{\mathrm{n}},\sum_n)}\label{equ9}
\end{equation}

2) M step: estimate new $\omega_{k}$, $\mu_{k}$, $\sigma_{k}$:
\begin{equation}
      \boldsymbol{\mu}_{\mathrm{k}}^{\text {new }}=\frac{1}{\mathrm{~N}_{\mathrm{k}}} \sum_{\mathrm{n}=1}^{\mathrm{N}} \gamma\left(\mathrm{z}_{\mathrm{nk}}\right) \boldsymbol{x}_{\mathrm{n}}\label{equ10}
\end{equation}

\begin{equation}
      \boldsymbol{\Sigma}_{\mathrm{k}}^{\text {new }}=\frac{1}{\mathrm{~N}_{\mathrm{k}}} \sum_{\mathrm{n}=1}^{\mathrm{N}} \gamma\left(\mathrm{z}_{\mathrm{nk}}\right)\left(\boldsymbol{x}_{\mathrm{n}}-\boldsymbol{\mu}_{\mathrm{k}}^{\text {new }}\right)\left(\boldsymbol{x}_{\mathrm{n}}-\boldsymbol{\mu}_{\mathrm{k}}^{\text {new }}\right)^{\mathrm{T}} \label{equ11}
\end{equation}

\begin{equation}
      \omega_{\mathrm{k}}^{\text {new }}=\frac{\mathrm{N}_{\mathrm{k}}}{\mathrm{N}} \label{equ12}
\end{equation}
where
\begin{equation}
      \mathrm{N}_{\mathrm{k}}=\sum_{\mathrm{n}=1}^{\mathrm{N}} \gamma\left(\mathrm{z}_{\mathrm{nk}}\right). \label{equ13}
\end{equation}\par
The stopping criterion for iterations is that the likelihood value between two iterations is less than $\tau$.

\section{Method}\label{sec4}
\subsection{Overview}\label{subsec2}

\begin{figure*}[htb]
  \centering
  \mbox{}
  \includegraphics[width=\linewidth]{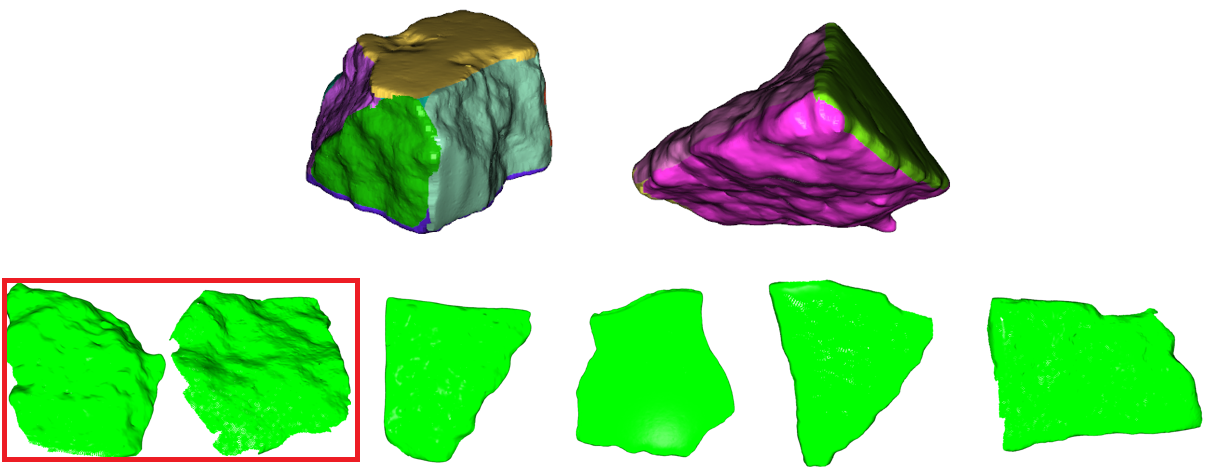}
  \mbox{}
  \caption{\label{fig:ex2}
         The illustration of the segmentation results and the orientation of the fractured surfaces.}
\end{figure*}

Our proposed method enables the automatic matching of fragments, which can be aligned for reconstructing broken objects. Our reassembly process is conducted exclusively on the fractured surfaces obtained by segmenting the model using the integral invariance method described in \cite{fractured}. The segmentation results are presented in Figure \ref{fig:ex2}
(a). Additionally, it is important to note that the orientation of the fractured surfaces is identical, as illustrated in Figures \ref{fig:ex2}(b)-(c).

\begin{figure*}[tb]
  \centering
  \mbox{}
  \includegraphics[width=\linewidth]{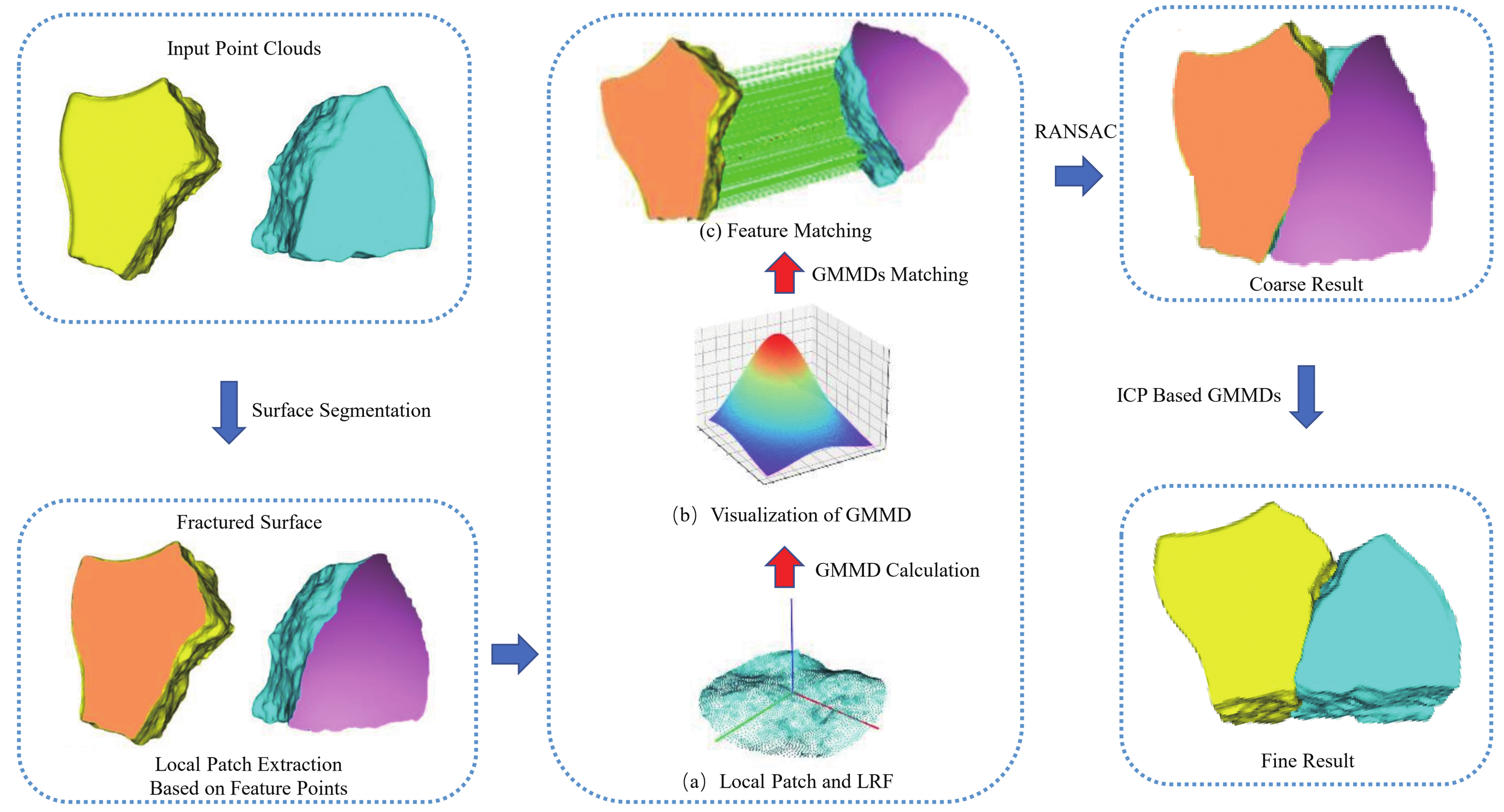}
  \mbox{}
  \caption{\label{fig:ex1}%
     The flow chart of GMD-based fragment reassembly method. 
     a) A local surface patch consisting of all points in the support region. 
     b) A downscaling of the 3D GMD into 2D for the purpose of visualizing the GMD. 
     c) GMD matching results where the green line represents a correct match and the red line represents an incorrect match.}
\end{figure*}

Figure~\ref{fig:ex1} illustrates the pipeline of our proposed method. First, we detect feature points on fractured surfaces using SIFT implemented in PCL, and the neighboring points within a sphere of radius $R$ centered on the feature points form a local surface patch on the fractured surface. Next, the surface patch is divided into different concave and convex regions based on the angle between the points and the local fitted plane. Given the concave and convex regions, we then create a regional GMD for each region, and these regional GMDs are merged to form the final GMD for the entire surface patch. Finally, matching fractured surfaces can be achieved by measuring the similarity between the GMDs.

\subsection{Feature extraction and LRF generation}\label{subsec3}
\label{FeatureLRF}

Given the fractured surfaces, we initially utilize the SIFT \cite{sift} method implemented in PCL to detect feature points on the fractured surfaces. Then the neighboring points within a sphere of radius $R$ centered on the feature points
form a local surface patch.

Next, we constructed the LRFs at these feature points to ensure the invariance of GMD to rotation and translation changes. Specifically, we calculate the covariance matrix of the points on the surface patch according to \cite{2010Unique}, the covariance matrix \textit{Cov} is defined as:

\begin{equation}
	\textit{Cov}=\frac{1}{N-1} \sum_{d_{p p_{i}}<R}\left(p_{i}-p\right)\left(p_{i}-p\right)^{T} \label{equ3}
\end{equation}

\noindent where $N$ is the number of points on the surface patch and $d_{p p_{i}}$ is the Euclidean Distance between points $p_{i}$ and $p$. 

Then the z-axis of LRF is computed as: 
\begin{equation}
	\overrightarrow{Z_{p}}=\left\{\begin{array}{cc}
\vec{n}_{p}, & \text { if }\enspace \left(\vec{n}_{p} \cdot \sum_{i=1}^{N} \overrightarrow{p_{i} p}\right)\geq 0 \\
-\vec{n}_{p}, & \text { otherwise }
\end{array}\right.  \label{equ4}
\end{equation}
where $\vec{n}_{p}$ is the eigenvector corresponding to the minimal eigenvalue of \textit{Cov} and $\overrightarrow{p_{i} p}$ is the vector between points $p_{i}$ and $p$. \par
Afterward, the x-axis of LRF is computed as:
\begin{equation}
	\overrightarrow{X_{p}}= \sum_{i=1}^{N} \overrightarrow{p_{i} p} \label{equ5}
\end{equation}

Finally, the y-axis of LRF can be computed as: 

\begin{equation}
      \overrightarrow{Y_{p}}=\overrightarrow{Z_{p}}\times \overrightarrow{X_{p}} \label{equ6}
\end{equation}

Given the LRF, the coordinates of all points on the local surface patch are converted into this LRF.

\subsection{GMD generation}\label{subsec4}

Given the local surface patches and their LRFs, in this stage, we use GMM to fit the local surface patches around the feature points, and then EM algorithm \cite{Embretson2000ItemRT} is used for estimating the parameters of GMM, thereby forming the descriptor GMD.

We consider the point distribution of the surface patch as a mixture of $k$ Gaussian distributions. Obviously, $k$ is an important parameter that should be set first. In previous work, the $k$ value of GMM generally equals to the number of points on the point cloud, for example in \cite{5674050}, which leads to large computational costs. Furthermore, an excessively large $k$ value makes GMM more sensitive to outliers and noise. 

Therefore, we divide the surface patch into concave and convex regions \cite{Son2017ReassemblyOF} and then use x-means \cite{xmeans} to calculate the $k$ value for GMD. The division of the concave and convex regions is the basis for using x-means to calculate $k$. For example, as shown in Figure~\ref{fig:ex5}(a), due to the continuity of the surface, x-means considers the entire surface as one cluster, whereas in Figure~\ref{fig:ex5}(b), concave regions (red parts) are divided into two clusters and convex region (green parts) are considered one cluster. Afterward, regional GMD is generated for concave and convex regions, respectively, and the final GMD is formed by merging the regional GMDs.

\subsubsection{Regional GMD generation}\label{subsubsec2}
For generating the regional GMDs, we first classify the points on the surface patch into concave region and convex region, by comparing the normal vector of the points and the the normal vector of the plane, which is fitted by the edge points of the surface patch. The direction of the normal vector of the fitted plane is specified as the same direction as the z-axis of the LRF. Then an arbitrary point can be classified using Equ. (\ref{equ8}).

\begin{equation}
      \operatorname{Con}\left(p_{i}\right)=\left\{\begin{array}{cc}
1, & \text { if }\enspace\left(\vec{n_{p_i}} \cdot \vec{n_{plane}}\right) \geq 0 \\
-1, & \text { otherwise }
\end{array}\right. \label{equ8}
\end{equation}
where $\vec{n_{i}}$ is the normal vector of the point and $\vec{n_{plane }}$ is the normal vector of the fitted plane. 1 means the point is in a convex region, whereas -1 means the point is in a concave region.

\begin{figure}[htb]
  \centering
  \mbox{}
  \includegraphics[width=.6\linewidth]{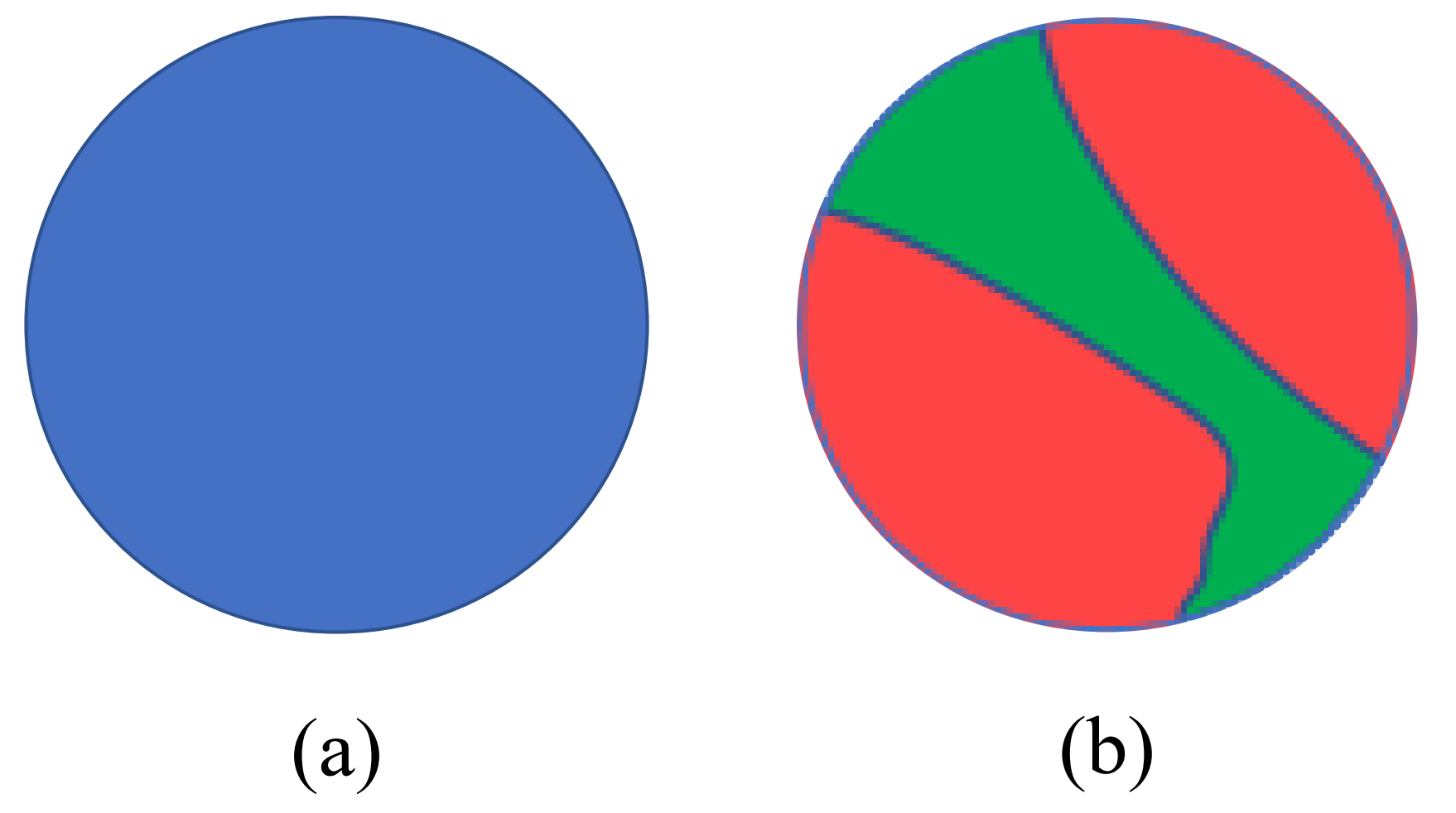}
  \mbox{}
  \caption{\label{fig:ex5}%
The illustration of the concave
and convex regions on a local surface patch. (a) The local surface patch; (b) The resulted concave and convex regions.}
\end{figure}

Given the concave and convex regions, x-means \cite{xmeans} is used for computing the $k$ values for concave and convex regions. Particularly, $k_{1}$ and $k_{2}$ represent the $k$ values of the concave and convex regions, respectively. In the example shown in Figure~\ref{fig:ex5}, there are two concave regions and one convex region, and $k_1=1$ for the convex region and $k_2=2$ for concave regions are obtained.

Given the \textit{k} value, the EM algorithm is then performed to fit the regional GMD for the concave and convex regions, respectively. Specifically, the cluster centers obtained using x-means are used as the initial $\mu$, and the sample variances of each cluster are used as the initial $\Sigma$. The weights $\omega$ of each Gaussian distribution are then initialized to $\frac{E_{region_i}}{E}$, $E_{region_i}$ represents the points' number of $i$th region, and $E$ represents the number of 
all points on the surface patch.

\subsubsection{The GMD of the surface patch}\label{subsubsec3}
The GMD descriptor of the entire surface patch consists of a regional GMD of the convex region and a regional GMD of the concave region. Considering that different sizes of concave and convex regions contribute unequally to the shape of the surface patch, we set different weights for the concave GMD and convex GMD, to form the final GMD of the surface patch. 

In particular, we use $f_{conc}$ and $f_{conv}$ to represent the concave GMD and convex GMD, then, the GMD of the entire surface patch can be calculated using Equ.(\ref{equ14}):

\begin{equation}
      \mathit{f}=\frac{\mathit{E}_{\mathit{conc}}}{\mathit{E}}f_{\mathit{conc}}+\frac{\mathit{E}_{\mathit{conv}}}{\mathit{E}}f_{\mathit{conv}} \label{equ14}
\end{equation}
where $E_{conc}$ and $E_{conv}$ represent the number of points in the concave and convex regions and \textit{E} represents the number of all points on the surface patch.

To reduce the dimension of the descriptor, we use the GMM parameters instead of the GMM value of each point to form the GMD, that is:

\begin{equation}
f=\{[\omega_1,...,\omega_k],[\mu_1,...,\mu_k],[\Sigma_1,...,\Sigma_k]\}
\end{equation}

\noindent where $\omega_{i}$, $\mu_{i}$ and $\Sigma_{i}$ are the parameters of the $k$th Gaussian distribution. 
The dimension of the GMD is $k+(3 \times k)+(3 \times 3\times k)$.

\subsection{GMD matching}\label{subsec5}
The matching between two fragments can be regarded as the GMD matching of two fractured surfaces. Therefore, in this stage, we measure the similarity of the GMDs.

Assume that we have a source fractured surface $\mathbf{P}^{s}$ and a target fractured surface $\mathbf{P}^{t}$, each contains a set of feature points $\left\{\mathbf{P}_{1}^{s}, \mathbf{P}_{2}^{s}, \cdots, \mathbf{P}_{n}^{s}\right\}$ and $\left\{\mathbf{P}_{1}^{t}, \mathbf{P}_{2}^{t}, \cdots, \mathbf{P}_{m}^{t}\right\}$, with their GMDs $\left\{\mathbf{f}_{1}^{s}, \mathbf{f}_{2}^{s}, \cdots, \mathbf{f}_{n}^{t}\right\}$ and $\left\{\mathbf{f}_{1}^{t}, \mathbf{f}_{2}^{t}, \cdots, \mathbf{f}_{m}^{t}\right\}$, respectively. 
 The $L_{2}$ distance is used to measure the similarity between the GMDs, which is defined as:

\begin{equation}\begin{aligned}d_{L_{2}}(f_{i}^{s},f_{j}^{t})&=\sum_{k=l}^{k_{s}}\sum_{l=l}^{k_{s}}\omega_{ik}^{s}\omega_{il}^{s}f(O\mid\mu_{ik}^{s}-\mu_{il}^{s},\Sigma_{s}+\Sigma_{s})\\&+\sum_{k=l}^{k_{t}}\sum_{l=l}^{k_{t}}\omega_{jk}^{t}\omega_{jl}^{t}f(O\mid\mu_{ik}^{t}-\mu_{il}^{t},\Sigma_{t}+\Sigma_{t})\\&-2\times\sum_{k=l}^{k_{s}}\sum_{l=l}^{k_{t}}\omega_{ik}^{s}\omega_{jl}^{t}f(O\mid\mu_{ik}^{s}-\mu_{il}^{t},\Sigma_{s}+\Sigma_{t}).\end{aligned}
\label{equ15}
\end{equation}

Two feature points are validated as matched correspondences only if the $L_{2}$ distance between their GMDs is small enough. The corresponding point pairs achieved through the matching of GMDs are fed into RANSAC \cite{FISCHLER1987726} and ICP \cite{Besl1992AMF} to 
estimate the transformation matrix that aligns the matched fractured surfaces.

\section{Experimental results}
\subsection{Experimental setting}

All experiments are performed on an AMD Ryzen 7 4800H 2.9-GHz CPU with 16.0GB of RAM, running 64-bit Microsoft Windows 10, Visual Studio 2019, and PCL 1.11.1. 

\begin{figure}[b]
  \centering
  \mbox{}
  \includegraphics[width=\linewidth]{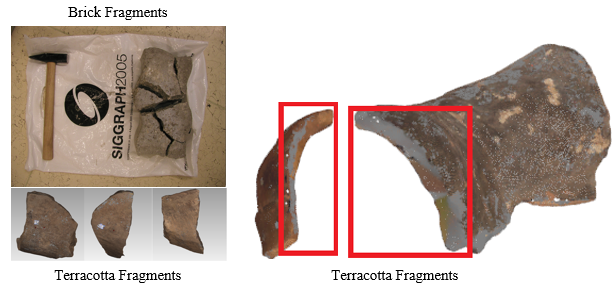}
  \mbox{}
  \caption{\label{fig:ex7}%
     Test models used in this work.}
\end{figure}

\subsection{Datasets}

We employed models of real-world broken objects with varying characteristics to validate the performance of our algorithm, as presented in Figure \ref{fig:ex7}. The point clouds of these broken objects were generated via 3D laser scanning \cite{10.1145/1141911.1141925}.

Specifically, the brick fragments comprised 6 pieces, and the fractured surfaces between the brick fragments are nearly planar, each with perfectly matched fractured surfaces. We use the brick fragments to evaluated the performance of the proposed method since it is a publicly available dataset. Then we apply the proposed method to the Terracotta fragment dataset aquired using laser scanner, which is also a real-scanned dataset and contains both thin and narrow fractured surfaces(marked in red boxes). 

The detailed information of the test data is listed in Table~\ref{table:t1} (in supplementary materials), where \textit{A} and \textit{B} represent a pair of fragments. 
Tests 1-3 are the fragments from the brick model, and Test 4 is the fragments of the Terracotta.

\subsection{Evaluation metrics}

Supposing $\mathbf{P}^{s}$ and $\mathbf{P}^{t}$ represent the source fractured surface and the target fractured surface, respectively. The feature points are extracted using the method in Section \ref{FeatureLRF}, and the GMD is calculated to describe local patches on $\mathbf{P}^{s}$ and $\mathbf{P}^{t}$. 
The distance between the GMDs of $\mathbf{P}^{s}$ and $\mathbf{P}^{t}$ is calculated using Equ.\eqref{equ15}. Then we use 6 evaluation indicators to evaluate the pairwise matching of two fragments, including:

1) \textbf{Percentage of Coverage (PoC)} is used to evaluate the completeness of the matching result of two fractured surfaces. To compute PoC, we project two fractured surfaces to one common plane $P_{c}$ that is fitted by the points on the bigger fractured surfaces. Specifically, $P_{c}$ is defined as: $A x+B y+C z=0$, where $\{A, B, C\}$ is the normal vector of the bigger fractured surfaces. Then all points on the two fractured surfaces can be projected to the plane using Equ.\ref{equ16}:

\begin{equation}
      \left\{\begin{array}{l}
			x=x_{i}-A t \\
			y=y_{i}-B t \\
			z=z_{i}-C t
			\end{array}\right.\label{equ16} 
\end{equation}
where
\begin{equation}
      t=\frac{A x_{i}+B y_{i}+C z_{i}+D}{A^{2}+B^{2}+C^{2}}. \label{equ17} 
\end{equation}

Supposing $\mathbf{P}^{s}$ is the bigger one with the the projected points $\mathbf{P}^{s'}$, and $\mathbf{P}^{t}$ is the smaller one with the projected points $\mathbf{P}^{t'}$. Then we compute the corresponding point pairs. Particularly, for each point $\mathbf{p}^{t'}_{i}$ in  $\mathbf{P}^{t'}$, if there exists any point $\mathbf{p}^{s'}_{j}$ in $\mathbf{P}^{s'}$ meets $L_2({P}^{t'}_{i}-{P}^{s'}_{j}) < \chi$, $\mathbf{p}^{t'}_{i}$ and $\mathbf{p}^{s'}_{j}$ are considered point pairs. Therefore, $PoC$ can be computed using: 

\begin{equation}
      PoC = \frac{|{P}^{o}|}{|{P}^{t'}|}. \label{equ18} 
\end{equation}

\noindent where $|P^{o}|$ is the number of point pairs and $|{P}^{t'}|$ is the number of points on $\mathbf{P}^{t'}$.

2) \textbf{Angle of the normal vector of the planes (AoNV)} is defined as the angle between the normal vectors of the fitted planes of the two fractured surfaces. The larger the angle between the normal vectors of the fitted planes, the greater the intersection formed by the two fragments. 

However, the plane fitted by all points on the fractured surface may tend to be closer to regions with higher concavity or convexity due to the fact that points in higher concave or convex regions contribute more. Therefore, we use edge points
of the fractured surface instead, which also reduces the impact of noise and better captures the overall spatial position of the fractured surface. 

3) \textbf{Angle of the normal vector of the local patch (localAoNV)} is defined as the angle between the normal vectors of the fitted planes of local surface patches. Since both PoC and AoNV fail to handle the case where the aligned fragments are flipped vertically, we propose localAoNV to evaluate the reassembly results. Although some studies have attempted to assess the quality of point cloud registration by measuring the cross-distance between paired fragments \cite{10.1007/s11263-022-01635-3}, we observed that the problem remains unresolved. 

For computing localAoNV, we introduced a "bounding box" to generate the local surface patch. To be more specific, for each feature point on $\mathbf{P}^{t}$, we generate a cube (denoted bounding box) with side length $l$, the patches of $\mathbf{P}^{s}$ and $\mathbf{P}^{t}$ locating in the bounding box, are considered the local surfaces patches used for computing localAoNV.
Then the localAoNV is computed in the same way as AoNV.

4) \textbf{Max Angle, Min Angle, and Mean Angle (MaA, MiA, MeA)} between normal vectors of all points on one fractured surface and its closest point on another fractured surface.

The larger the PoC and the smaller the AoNV, localAoNV, MaA, MiA, and MeA, the better the registration results.

\subsection{Parameter analysis}

\begin{figure*}[hbt]
  \centering
  \mbox{}
  \includegraphics[width=\linewidth]{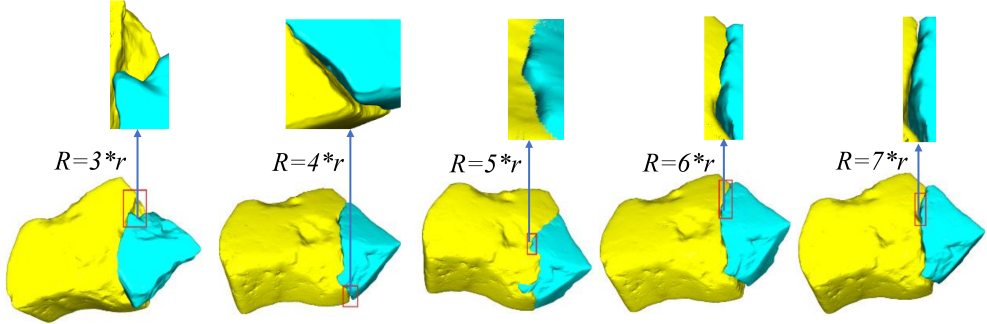}
  \mbox{}
  \caption{\label{fig:ex10}%
    The results of our method under different $R$ settings.}
\end{figure*}

The support radius $R$, which determines the scale of the surface patches to be described and matched, is an important parameter of the proposed method. Different $R$ can have a significant impact on the matching performance. For example, a large $R$ may result in a higher sensitivity to noise, while a small $R$ may induce insufficient points to calculate the GMD, leading to a failure in the matching process. Moreover, setting an extremely large $R$ may cause overlapping of two local surface patches that are close to each other, 
resulting in high similarity and matching failure. To determine the optimal parameter setting, we conduct a thorough analysis of the GMD's performance under various $R$ settings.

Figure~\ref{fig:ex10} presents the results of our proposed method using different \textit{R} and Table~\ref{table:t2} lists the scores of the GMD achieved under different $R$. The test models are brick models. We only change the values of $R$, which are $R=3*r$, $R=4*r$, $R=5*r$, $R=6*r$ and $R=7*r$ respectively. It can be found that the best performance is achieved when $R=6*r$, where $r$ is the point cloud resolution, which is calculated using Equ.\ref{equ19}:
\begin{equation}
      r = \frac{1}{M} \sum_{i=1}^{M} \min_{j \neq i} | p_i - p_j | \label{equ19}  
\end{equation}
where \textit{M} represents the size of the point cloud, $|\cdot |$ represents Euclidean Distance, and $\min_{j \neq i}$ means point $p_j$ is closest to point $p_i$.

\begin{table*}[t]
      \centering
      \caption{\label{table:t2}%
            The results under different $R$ settings and the best scores are in bold.}
      \small
      \setlength{\tabcolsep}{4mm}{
      \begin{tabular}{llllll}
      \hline
      \textit{R}  & 3*\textit{r}    & 4*\textit{r}   & 5*\textit{r}   & 6*\textit{r}    & 7*\textit{r}         \\ \hline
      PoC     & 89.13\% & 98.419\% & 98.419\% & \textbf{98.913\%} & 95.257\% \\
       AoNV      & 12.406   & 22.268  & 21.208  & \textbf{15.25} & 20.614    \\
      localAoNV  &99.08    & 25.071  &30.231  &\textbf{24.942}  &35.05   \\
      MaA  & 41.785   & 60.688  & 53.273  & \textbf{31.814}   & 34.444   \\
      MiA  & 0.994  & 0.645 & 0.638 & 1.39  & \textbf{0.296}   \\
      MeA & 14.604   & 21.86    & 25.947  & \textbf{11.564}  & 15.167   \\ 
      \hline
      \end{tabular}}
\end{table*}\par

\begin{table*}[]
      \centering
      \small
      \caption{\label{table:t3}%
            The scores of matching pairs with different methods. (Best scores are in bold and time is in seconds.)}
      \setlength{\tabcolsep}{3.5mm}{
      \begin{tabular}{lllllll}
      \hline
            & GMD     & TEASER   & GROR      & FPFH      & SHOT      & SI        \\ \hline
      PoC     & \textbf{98.913\%} & 91.996\% & 85.376\% & 95.257\% & 90.02\% & 97.233\% \\
    AoNV      & \textbf{15.2} & 28.118  & 163.005  & 175.455   & 168.46 & 16.9  \\
      localAoNV & \textbf{24.943} &91.21    &104.681      &148.549    &146.909    &30.024    \\
      MaA  & \textbf{31.814}   & 62.661  & 59.355   & 69.009   & 55.395   & 48.409   \\
      MiA  & 1.391  & 3.730  & 0.267  & \textbf{0.263}  & 0.439  & 0.808  \\
      MeA & \textbf{11.564}  & 28.986  & 17.386   & 14.266   & 14.266   & 24.82   \\ 
      time    & \textbf{1.029}	 & 2.770	& 1.366	& 2.872	& 3.814	& 1.471    \\ \hline
      \end{tabular}}
\end{table*}

\begin{figure*}[htb]
  \centering
  \mbox{}
  \includegraphics[width=\linewidth]{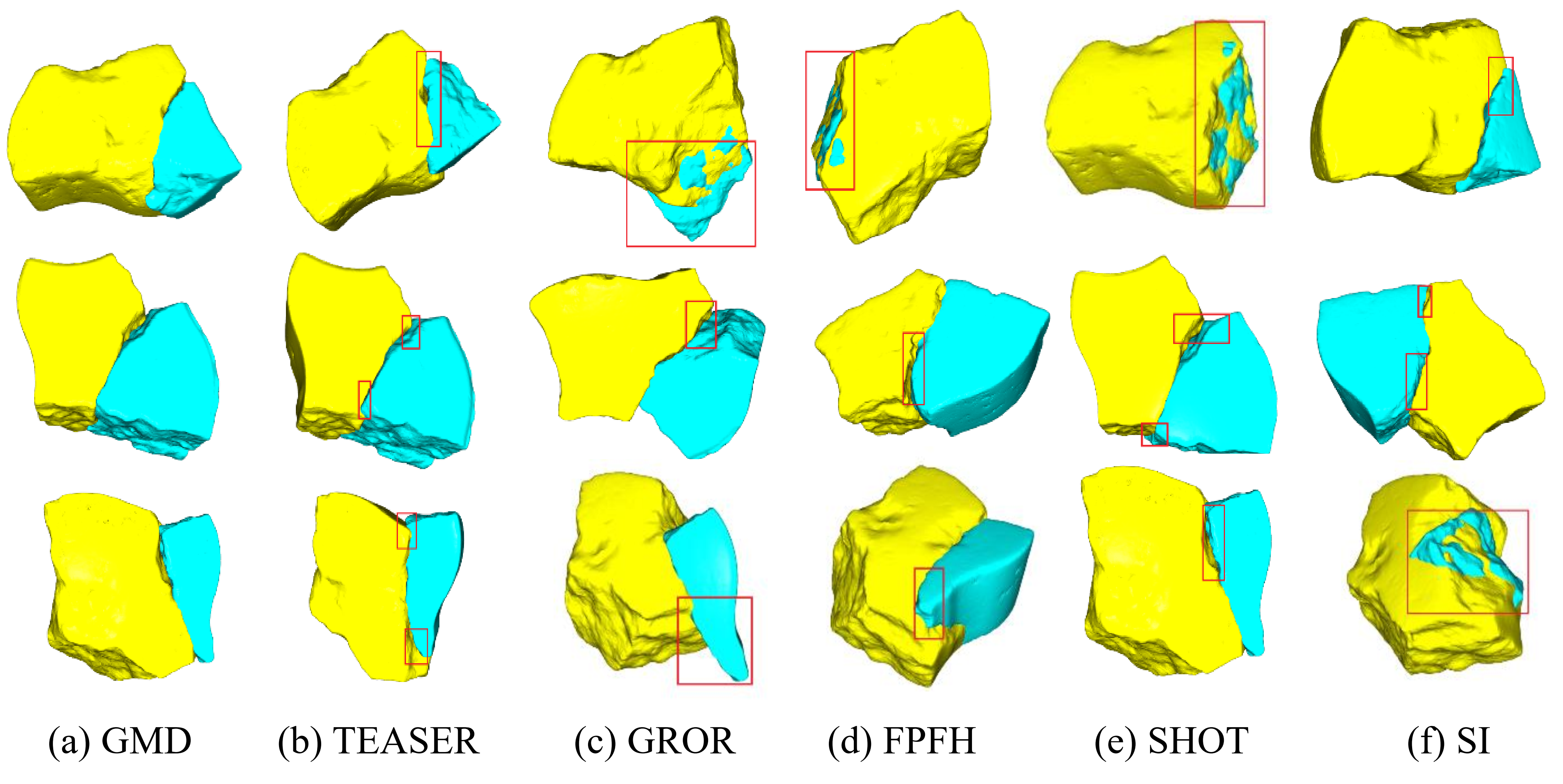}
  \mbox{}
  \caption{\label{fig:ex11}%
    The results of matching pairs from the Terracotta dataset were achieved using different methods.}
\end{figure*}

\subsection{Comparison results}

To further evaluate the effectiveness of our proposed algorithm, we conducted a comparison with existing methods such as TEASER \cite{Yang20tro-teaser} and GROR \cite{GROR}. 
Moreover, we also compared our method with commonly used feature descriptors, including FPFH \cite{rusu2009fast}, 
SHOT \cite{2014SHOT}, and Spin Image \cite{assfalg2007content}. 

We carried out tests on the fragments of the brick model and the scores are presented in Table~\ref{table:t3}. As can be found in Table~\ref{table:t3}, our GMD exhibits better performance compared to the other descriptors and methods. Furthermore, Figure~\ref{fig:ex11} also presents the results of the pairwise matching results of our method and the baselines. It can be observed that the proposed approach demonstrates a significant advantage over the other methods.
It can also be observed from table~\ref{table:t3} that the running time of our method is also superior to other methods.

\subsection{Robustness test}

\subsubsection{Gaussian Noise}
\begin{figure*}[htb]
  \centering
  \includegraphics[width=\linewidth]{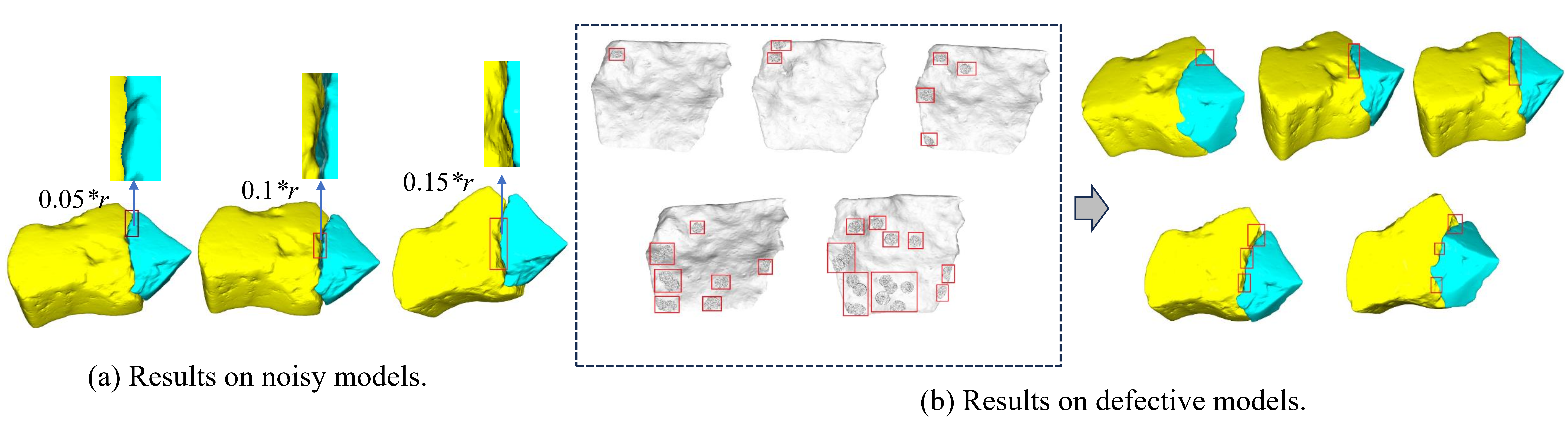}
  \caption{\label{fig:ex14}%
  The results on models with different sizes of noise and defective parts.}
\end{figure*}

To assess the robustness of the GMD to noises, we conducted experiments by applying Gaussian noise to each fractured surface with different standard variances $\sigma$, although the models have been perturbed by real noise induced during the scanning process. Specifically, we set standard variance $\sigma$ to $0.05*r$, $0.1*r$, and $0.15*r$. The results are presented in Table~\ref{table:t4} and Figure~\ref{fig:ex14} shows the results on models with different levels of noise. 

\begin{table}[htb]
      \caption{\label{table:t4}%
           The scores on models perturbed by different standard deviations of noises.}
           \small
      \setlength{\tabcolsep}{9mm}{
      \begin{tabular}{llll}
      \hline
      $\sigma$       & 0.5*\textit{r}       & 1.0*\textit{r}       & 1.5*\textit{r}      \\ \hline
      PoC     & \ 92.787\% & 91.403\% & 89.166\% \\
      AoNV      & 16.141  & \ 0.343  & 2.435  \\
      localAoNV & \ 33.282   &35.955    &42.5681 \\
      MaA  & 37.81     & 47.792   & 42.557  \\
      MiA  & 0.82  & 1.372   &\ 0.388 \\
      MeA & 16.392    & 15.327   & \ 11.65  \\ \hline
      \end{tabular}
      } 
\end{table}

As can be observed, the proposed GMD can maintain a relatively high matching accuracy even under noisy conditions. These results also demonstrate that the proposed GMD feature descriptor can effectively capture the geometric characteristics of fractured surfaces and can provide reliable and robust matching results even in the presence of noise.

\subsubsection{Robustness to abrasion}
In order to comprehensively assess the robustness of GMD, we also test our algorithm on real-scanned models. Furthermore, models with defective parts should also be taken into account. Both the real noise induced by the scanning process and the loss of parts disrupts the original geometry, presenting more difficulty for matching the fractured surfaces. 

Therefore, we randomly distort the local shape of some real-scanned fractured surfaces, as shown in Figure~\ref{fig:ex14}. Table~\ref{table:abrasion} and Figure~\ref{fig:ex14} show the results and as can be observed, GMD is also robust to small amounts of defective parts. However, the performance of GMD gradually decreases as the number of defective parts increases, in the 5th experiment in particular, we observed that when the number of defective parts exceeded 20, the resulted experimental outcomes were no longer satisfactory. This suggests that the fractured surface-based approach may not be effective when the shape of the fractured surface is significantly altered by damage.

\begin{table*}[htb]
\caption{\label{table:abrasion} The results on models with abrasion.}
\small
\setlength{\tabcolsep}{4mm}
{
    \begin{tabular}{lllllll}
    \hline
    Models & PoC & AoNV & localAoNV & MaA & MiA & MeA\\
    
    \hline
    No.1 & 97.179\% & 10.936 & 30.487 & 35.58 & 0.77 & 16.325 \\
    
    No.2 & 88.531\% & 12.741 & 31.911 & 48.021 & 0.288 & 20.415 \\
    
    No.3 & 85.51\%  & 14.306 & 33.795 & 65.982 & 0.175 & 25.341 \\
    
    No.4 & 83.452\% & 15.336 & 59.049 & 73.69  & 11.205 & 34.901 \\
    
    No.5 & 73.154\% & 16.612 & 72.563 & 48.783 & 0.033  & 45.869 \\
    \hline
    \end{tabular}
}
\end{table*}

\subsection{Ablation studies}

In this section, we conducted ablation experiments to demonstrate the advantage of feature points and GMD in the fragment reassembly framework.

\begin{figure}[htb]
  \centering
  \includegraphics[width=\linewidth]{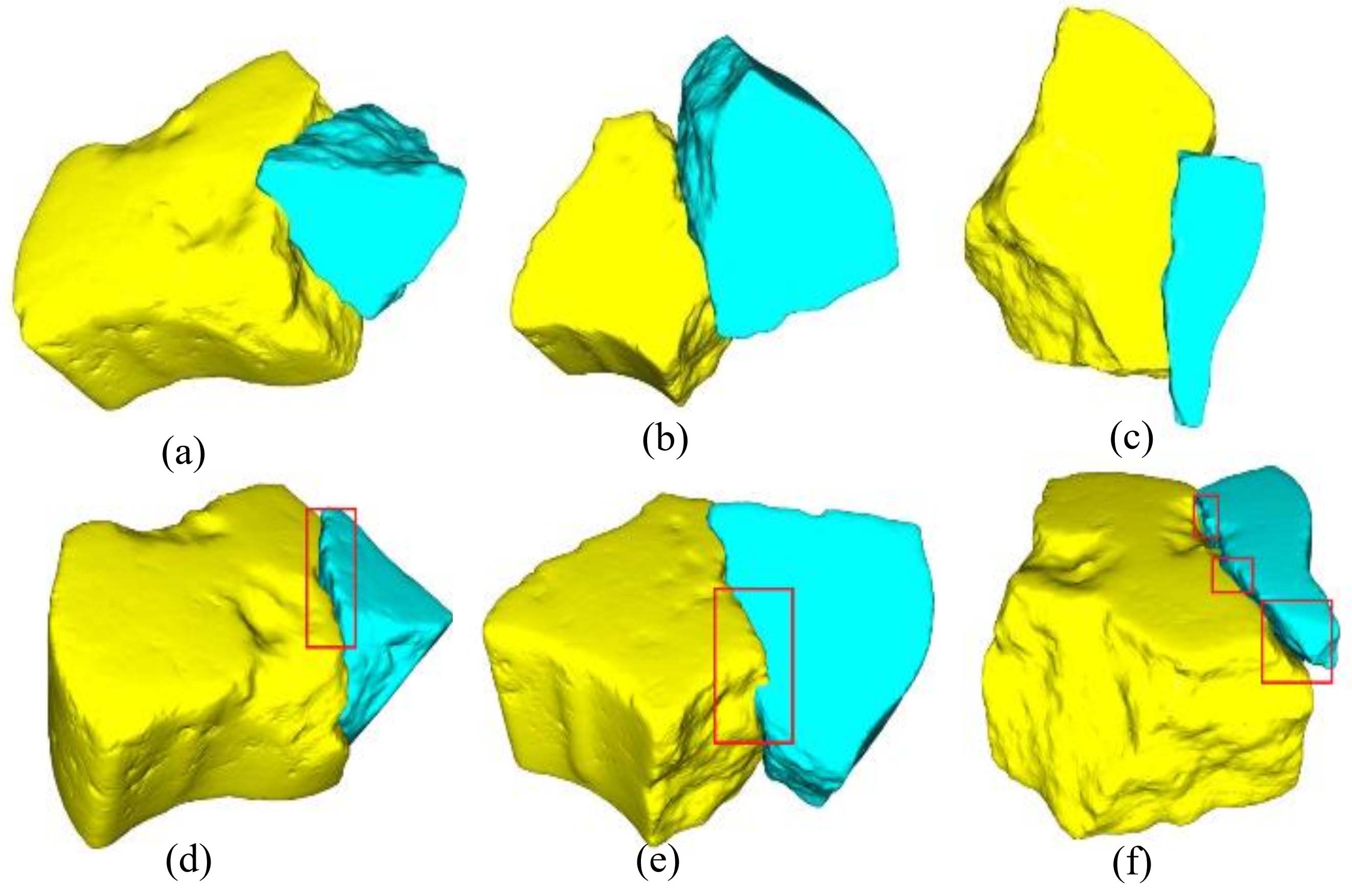}
  \caption{\label{fig:a2}
  The results of ablation studies.}
\end{figure}

To demonstrate the effectiveness of GMD, we directly used the feature points to align the fractured surface. To make a comparison, we also align the fractured surfaces based on GMD. Figure~\ref{fig:a2}(a$ \sim $c) shows the results of the experiments without GMD. We can find that in the absence of GMD, the final experimental results were not satisfactory. Consequently, all results basically showed deviations, as presented by Table~\ref{table:t6}.

\begin{table}[htb]
\caption{\label{table:t6} The results of ablation studies.}
\small
    \begin{tabular}{llllllll}
    \hline
    Ablation study & Models & PoC & AoNV & localAoNV & MaA & MiA & MeA\\
    \hline
    \multirow{3}{*}{w/o GMD} & No.1 & 78.557\% & 12.736 & 77.510 & 51.814 & 9.029 & 34.074 \\

    & No.2 & 57.859\% & 8.641 & 75.631 & 89.947 & 0.13 & 17.416 \\
    & No.3 & 71.624\% & 19.91 & 84.506 & 64.296 & 5.444 & 36.032 \\
    \hline
    
    \multirow{3}{*}{defective surfaces} & No.1 & 92.764\% & 10.375 & 25.732 & 49.571 & 0.035 & 17.864 \\
    & No.2 & 93.393\% & 9.426 & 29.95 & 65.434 & 0.095 & 20.9\\
    & No.3 & 80.312\% & 6.310 & 26.663 & 60.153 & 0.089 & 20.88\\
    \hline
           
    \end{tabular}
\end{table}

Furthermore, we also validate the necessity of feature points. As shown in Figure~\ref{fig:a2}(d$ \sim $f) and Table~\ref{table:t6}, we can see that without the support of feature points, the final experimental results also exhibit some deviation. This is because the neighboring points are too close to each other and thus the two neighboring GMDs overlap too much.

\subsection{Application on Terracotta fragments}
\begin{figure*}[htb]
  \centering
  \includegraphics[width=\linewidth]{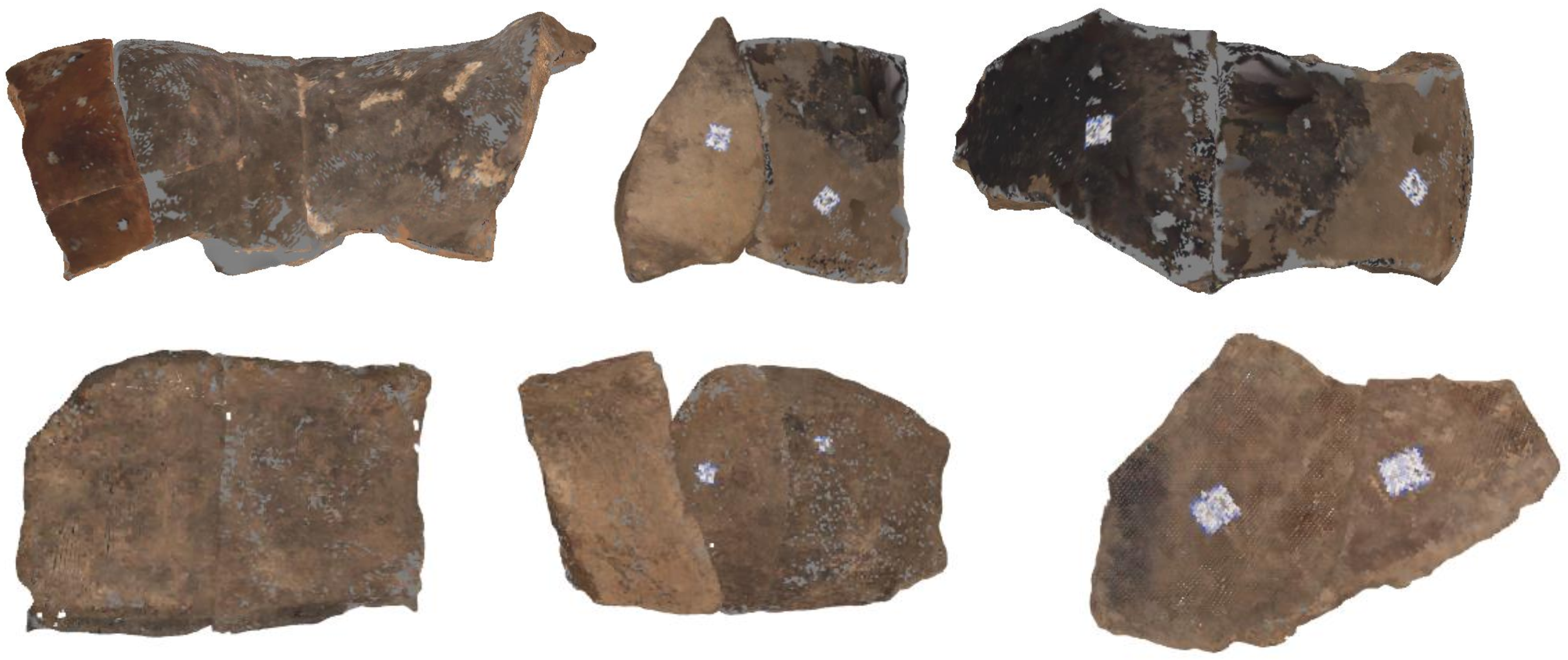}
  \caption{\label{fig:ex18}%
    The results of pairwise matching of Terracotta fragments.}
\end{figure*}

We applied our method on real-scanned Terracotta fragments which have complex shapes and narrow fractured surfaces, as shown in Figure~\ref{fig:ex18}. The satisfactory results still validate the effectiveness of the proposed method on narrow fractured surfaces and partially matched fractured surfaces, highlighting the utility of the proposed method.

\subsection{Limitation}

The proposed method struggles with fractured surfaces that are flat and lack distinctive feature points, a limitation common to many feature descriptors. Additionally, our method has difficulty handling cases where two fractured surfaces have significantly different sampling densities. Specifically, when one surface undergoes substantial downsampling—especially with a rate exceeding 60\%—while the other retains its original resolution, the method is prone to failure.

\section{Conclusions}
This work proposes a novel GMD and an algorithm for reassembly of fragments. In this approach, two fragments are considered adjacent if their GMDs of the fractured surfaces matched. We have conducted thorough tests on the proposed algorithm using both publicly available dataset and our Terracotta dataset. Furthermore, to quantitatively evaluate the performance of our method, we also proposed six evaluation metrics. The experimental results demonstrate that the GMD-based reassembly algorithm outperforms several commonly used feature descriptors 
and recently proposed methods in terms of accuracy and robustness, indicating the promising potential of GMD for practical applications on fragment reassembly and cultural heritage reconstruction.

However, our proposed method still has some limitations. In particular, our method may fail when the density difference between two given point clouds is too significant. We observed that our method becomes unfeasible when down-sampling one of the point cloud data while retaining the other data at its original resolution, particularly when the down-sampling rate is more than 60\%, due to Equ.(\ref{equ14}).  To address these limitation, our future research efforts will be directed toward investigating and addressing the impact of point cloud density on our approach.

In addition to the previously mentioned six evaluation metrics, we also employed a commonly used metric, \textit{i.e}., RMSE (Root Mean Square Error), which measures the error between corresponding point pairs in terms of Euclidean Distance, to assess the performance of our matching results. Furthermore, we also used another commonly used metric PRC to evaluate the performance of all selected descriptors. The precision is the number of correct matches concerning the total number of matches, and the recall is the number of correct matches with respect to the number of corresponding features.


    
    
    

\clearpage

\begin{appendices}

\section{Supplemental document}\label{secA1}

\setcounter{table}{0}
\renewcommand{\thetable}{A\arabic{table}}
\renewcommand*{\theHtable}{\thetable}

Algorithm~\ref{alg::conjugateGradient} summarizes our method for reassembling fractured surfaces $\mathbf{P}^{s}$ and $\mathbf{P}^{t}$, presented in pseudocode form.

\begin{algorithm}
		\renewcommand{\algorithmicrequire}{\textbf{Input:}}
		\renewcommand{\algorithmicensure}{\textbf{Output:}}
		\caption{\label{alg::conjugateGradient}%
                  Fragment Reassembly based on GMD.} 
		\begin{algorithmic}[1]  
			\REQUIRE
			Two fractured surfaces $\mathbf{P}^{s}$ and $\mathbf{P}^{t}$.
			\ENSURE
			The best transform matrix between $\mathbf{P}^{s}$ and $\mathbf{P}^{t}$.
			\STATE $f_a$, $f_b$ = $SIFT\_detection(\mathbf{P}^{s},\mathbf{P}^{t})$
			\FOR{for $p_a$ in $f_a$}
            \STATE GenerateLRF($p_a$, $r$)
                \STATE $g_a$ =GenerateGMD()
				\FOR{for $p_b$ in $f_b$}
					\STATE $g_b$ =GenerateGMD()
					\STATE $d_L$=CaculateL2Distance($g_a$ , $g_b$);
					\IF{$d_{L}<_\zeta$}
						\STATE matchedPoints.add($p_a$,$p_b$);
                        \STATE matchedGMDs.add($g_a$ , $g_b$)
					\ENDIF
				\ENDFOR
			\ENDFOR
            \STATE $d_{AB}$ = CaculateL2Distance(matchedGMDs)
            \IF{$d_{AB}<_\psi$}
                \STATE RANSAC(matchedPoints);
                \STATE ICP($\mathbf{P}^{s}$, $\mathbf{P}^{t}$);
            \ENDIF
		\end{algorithmic} 
\end{algorithm}

The details of fragments are listed in Table~\ref{table:t1}.

\begin{table*}
\caption{\label{table:t1} The details of fragments.}
\centering
\setlength{\tabcolsep}{2mm} 
\small 
\begin{tabular*}{\textwidth}{@{\extracolsep\fill}lcccccc}
\toprule
& \multicolumn{2}{@{}c@{}}{\#Fractured surfaces} & \multicolumn{2}{@{}c@{}}{\#Points size} & \multicolumn{2}{@{}c@{}}{\#Feature Points} \\ \cmidrule{2-3}\cmidrule{4-5}\cmidrule{6-7}
Test & A & B & $\mathbf{P}^{s}$ & $\mathbf{P}^{t}$ & $\mathbf{P}^{s}$ & $\mathbf{P}^{t}$ \\
\midrule
1 & 8 & 6 & 48647 & 90420 & 586 & 600 \\
2 & 8 & 6 & 77415 & 80654 & 919 & 899 \\
3 & 4 & 8 & 60971 & 93593 & 745 & 1001 \\
4 & 4 & 6 & 3589 & 2670 & 65 & 69 \\
\bottomrule
\end{tabular*}
\end{table*}

In addition to the previously mentioned six evaluation metrics, we also employed a commonly used metric, \textit{i.e}., RMSE (Root Mean Square Error), which measures the error between corresponding point pairs in terms of Euclidean Distance, to assess the performance of our matching results. Furthermore, we also used another commonly used metric PRC to evaluate the performance of all selected descriptors. The precision is the number of correct matches concerning the total number of matches, and the recall is the number of correct matches with respect to the number of corresponding features. 

Figure~\ref{app:a1} shows the RMSE scores and PRC scores achieved by the proposed method and baselines, indicating the superior ability of our algorithm except for the RMSE in the z-axis. This is because the matching pairs achieved using SHOT and SI appear to be interpenetrated, as shown in Figure~\ref{fig:ex11}.

\setcounter{figure}{0}
\renewcommand{\thefigure}{A\arabic{figure}}
\renewcommand*{\theHfigure}{\thefigure}

\begin{figure}[htb]
  \centering
  \includegraphics[width=\linewidth]{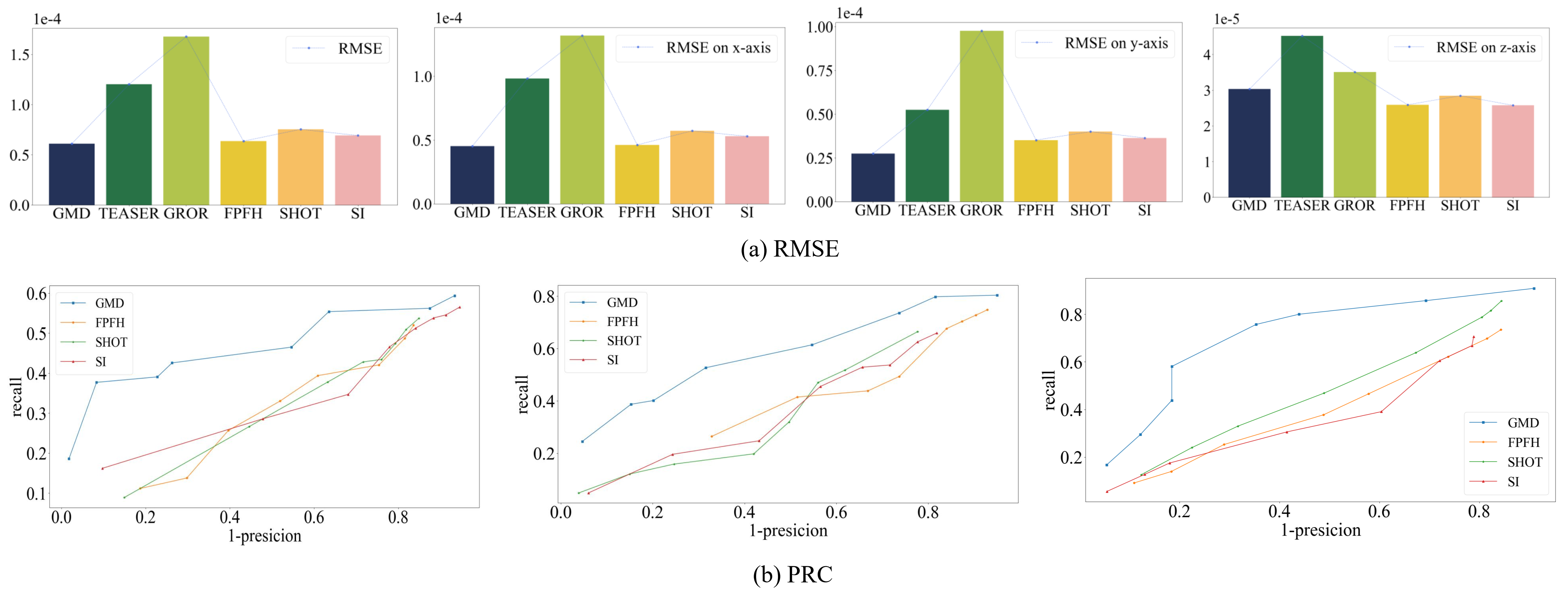}
  \caption{The results of RMSE and PRCs.}
  \label{app:a1}
\end{figure}




\end{appendices}


\clearpage

\bibliography{sample}

\end{document}